\documentclass[final,3p,times]{elsarticle}

\usepackage{epsfig}
\usepackage{amssymb}
\usepackage{amsmath}
\usepackage{amsthm}

\usepackage{subcaption}
\usepackage{lineno}
\usepackage{booktabs}
\usepackage{algorithm}
\usepackage{algorithmic}
\usepackage{amsmath}
\newtheorem{theorem}{Theorem}
\newtheorem{lemma}{Lemma}

\journal{Neurocomputing}

\begin{document}

\begin{frontmatter}



\title{Multi-view clustering integrating anchor attribute and structural information}


\author{Xuetong Li}
\ead{xuetongli@sjtu.edu.cn}
\author{Xiao-Dong Zhang\corref{cor1}}
\ead{ xiaodong@sjtu.edu.cn}

\cortext[cor1]{Corresponding author}
\affiliation{organization={School of Mathematical Sciences,Ministry of Education (MOE) Funded Key Lab of Scientific and Engineering Computing, Shanghai Center for Applied Mathematics},
            addressline={Shanghai Jiao Tong University}, 
            city={Shanghai},
            postcode={200240}, 
            country={P.R.China}}

\begin{abstract}
Multisource data has driven the development of advanced clustering algorithms, such as multi-view clustering, which critically rely on the construction of similarity matrices. Traditional algorithms typically generate these matrices based solely on node attributes. However, for certain directed real-world networks, neglecting the asymmetric structural relationships between nodes may compromise the accuracy of clustering results. This paper introduces a novel multi-view clustering algorithm, AAS, which employs a two-step proximity approach using anchors in each view, effectively integrating both attribute and directed structural information. This method enhances the clarity of cluster features within the similarity matrices. The construction of the anchor structural similarity matrix utilizes strongly connected components of directed graphs. The entire process—from the construction of similarity matrices to clustering—is formulated within a unified optimization framework. Comparative experiments conducted on the modified Attribute SBM dataset, benchmarked against seven other algorithms, demonstrate the effectiveness and superiority of AAS.
\end{abstract}




\end{frontmatter}



\section{Introduction}

The rapid development of informatization has led to a dramatic increase in the volume of data, making it particularly important to analyze these data to reveal their inherent features, identify potential structures and patterns. Clustering, as an unsupervised learning technique for finding the intrinsic structure of data, has a wide range of application prospects. In data mining, clustering can be used to provide a preprocessing function, standardizing complex structured multidimensional data \cite{garcia2015data}. In the business sector, clustering serves as an effective market segmentation tool. By analyzing consumer behavior and characterizing the features of different customer groups, potential markets can be discovered, thereby facilitating the operation of recommendation systems \cite{forouzandeh2023new}. In the realm of Internet applications, clustering can be applied to online document categorization and image segmentation in image processing \cite{duarte2023review,tang2022contrastive,zhang2022novel}. These applications demonstrate the widespread use and importance of clustering across multiple domains.

Clustering based solely on a single aspect of data samples can lead to biased results and does not align with the human tendency to view things from multiple perspectives \cite{shi2019semi}. Therefore, multi-view clustering algorithms have been developed to allow information from multiple views to complement each other and to uncover potential consensus structures. Lele Fu et al. \cite{fu2020overview} classified multi-view clustering algorithms into graph-based models \cite{li2015large}, space-learning models \cite{zhang2015low}, and binary code learning models \cite{zhang2018binary}, with the former two being more extensively studied. Graph-based clustering algorithms utilize the structure and properties of graphs for clustering, such as spectral clustering. Graph-based multi-view clustering focuses on the weighting of information from multiple views \cite{nie2016parameter}, improving spectral clustering steps \cite{nie2016constrained}, and employing methods such as anchors to reduce algorithm complexity \cite{kang2020large,cai2014large,liu2022scalable}, paying attention to the multi-order or local similarity of samples \cite{hao2020self,mei2023multi}, and dealing with incomplete data \cite{gu2024edison,wang2024joint,liu2024low}. Space-learning clustering, on the other hand, projects data into another space to achieve distinct clustering effects, focusing on reconstructing networks in the ideal space \cite{zhang2017latent}, combining consensus and complementary information from multiple views \cite{wang2017exclusivity}, and multi-step fusion optimization \cite{zhong2023self}.

\begin{figure}[H]
    \centering
    \includegraphics[width=1\linewidth]{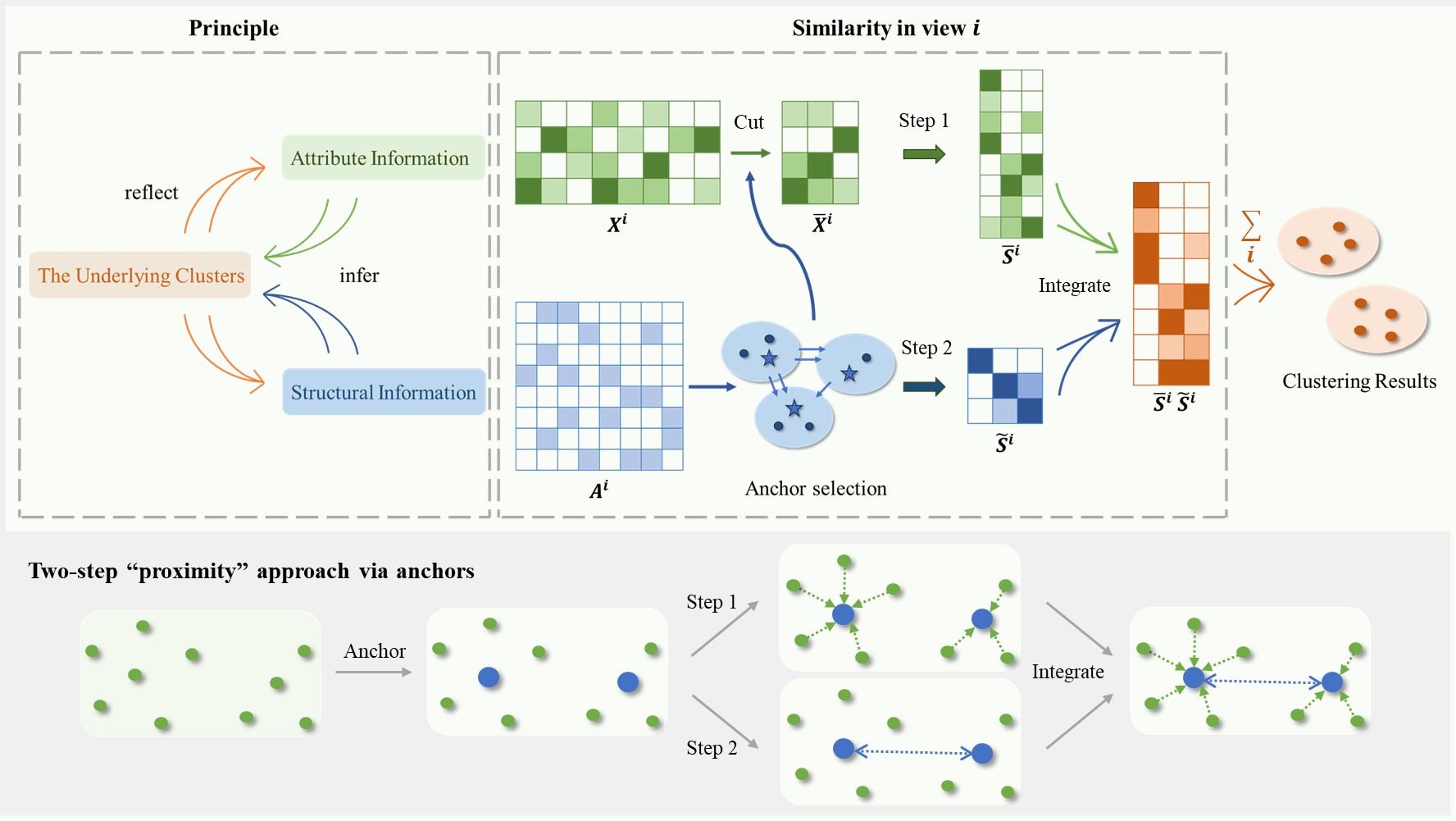}
    \caption[Short caption]{Overview of the AAS framework. The principle of the AAS and its matrix-based framework are shown on a green background. Green and blue matrices illustrate the computation of similarity matrices using attribute and structural information, respectively, which are then integrated for multi-view clustering. Anchor selection and the two-step proximity approach in the $\overline{S}^{i}\Tilde{S}^{i}$ process are visualized with nodes on a grey background. The first step reduces the distance between nodes and nearby anchors based on attribute information. The second step brings anchors within the same strongly connected components closer together based on structural information. "Proximity" is indicated by increased matrix values, aiming to produce a similarity matrix with clearer implicit cluster features.}
    \label{fig:Framework}
\end{figure}

While multi-view clustering algorithms have advanced significantly, most current research focuses on clustering based on node attributes, often neglecting structural information between nodes. In real-world scenarios like online social networks, biological neural networks, and computer communication networks, edge structure is crucial for node classification. Considering both attribute and structural information is essential for uncovering the network's underlying features, as illustrated in Fig.\ref{fig:Framework}. However, few studies address both node attributes and structure. For clustering on a single view, Kamal Berahmand et al. \cite{berahmand2022new} calculate the structural and attribute similarities separately and then linearly combine them. Mansoureh Naderipour et al. \cite{naderipour2022fuzzy} compute the attribute and structural similarities of samples separately before integrating them based on their similarity. For the multi-view setting, Yan W et al. \cite{yan2023collaborative} and Kong G et al. \cite{kong2024multi}also strive to integrate the structure between samples and the intrinsic attribute information. However, these works consider the abstract spatial structure based on sample attribute vectors, rather than the topological structure between entities, and the input consists solely of sample attributes. Khameneh A Z et al. \cite{khameneh2023multi} also adopt a linear combination to integrate attribute and structural similarities, uniquely utilizing directed graph structures but converting the asymmetric matrix into a symmetric one in subsequent processing to eliminate the direction of edges. 

In many real-world networks, relationships are directional, and ignoring this can distort the true clustering structure. For example, on online platforms, trust ratings between users are directed, and disregarding this may lead to incorrect identification of trust communities. In economic trade networks, import and export relationships are directed; treating them as undirected can misjudge trade surpluses or deficits, affecting the delineation of economic circles. Additionally, some specific structural patterns such as strongly connected components are valuable references for clustering in directed networks.

To reduce the complexity of clustering algorithms, some studies have adopted the anchor method. The basic steps and required proofs can be found in the literature \cite{kang2020large}. Similar methods have been applied in \cite{cai2014large,liu2022scalable}, differing in their selection of anchors. Some methods aggregate node degrees across multiple views, while others use K-means on each view to identify cluster centers. However, these studies commonly use a fixed number of anchors across multiple views, sometimes applying the same set of anchors directly to different views. This restricts anchor selection and may not ensure their representativeness for a specific view. Moreover, most current anchor-based methods handle multiple optimization tasks, like constructing the similarity matrix, graph partitioning, and discretizing continuous labels, separately, which can lead to suboptimal results.

This paper explores how to optimize the use of multisource data samples, specifically focusing on attribute and directed structural information for effective clustering. Our novel method for processing directed structural information sets our research apart from previous studies. The main contributions and innovations of this paper are:\\
\textbullet We propose a novel similarity matrix based on a two-step proximity approach, utilizing anchors as intermediaries within each view. Initially, the attribute similarity matrix increases the similarity between data nodes and their class-matching anchors. Subsequently, a structural similarity matrix augments similarity values among anchors within the same strongly connected component. Integrating this structural similarity as prior knowledge into the attribute similarity enhances the clarity of cluster features in the resulting matrix.\\
\textbullet We propose a novel strategy for selecting anchors, which identifies underlying clusters from the strongly connected components of the network, aiming to represent each cluster with appropriate anchors. The structural similarity matrix is then defined based on the relationships between anchors and their components, effectively bringing similar cluster anchors closer.\\
\textbullet During the anchor selection process based on directed structures, we introduce a new algorithm that simultaneously identifies strongly connected components in directed graphs and computes node centrality within these components. We also provide a necessary and sufficient condition for strong connectivity in directed graphs.\\
\textbullet We have adjusted the existing NESE algorithm and the Attribute SBM data generation method to meet the requirements of this study. By integrating the entire clustering process into a unified optimization framework and conducting comparative simulation experiments, we demonstrate the AAS algorithm's effectiveness.

In Section 2, we review the multi-view subspace clustering methods, anchor methods, and multi-view spectral clustering via integrating nonnegative embedding and spectral embedding (NESE). Section 3 describes the proposed algorithm, abbreviated as AAS, according to the steps of constructing the anchor structure similarity matrix and the fusion optimization framework. Section 4 presents the optimization algorithm for AAS and analyzes its time complexity and convergence. Section 5 generates multi-view networks compatible with the proposed algorithm and reports the experimental results of AAS, comparing it with other algorithms. Additionally, it analyzes the runtime, convergence, ablation studies, and parameter sensitivity.Section 6 concludes the paper.

\section{Preliminaries}
\subsection{Notations}

\begin{table}[H]
\centering
\begin{tabular}{ll}
\toprule
Notation & Description \\
\midrule
$n$ & Number of samples \\
$k$ & Number of clusters \\
$d_{i}$ & Dimension of attributes in view $i$ \\
$m_{i}$ & Number of anchors in view $i$ \\
$X^{i}=[x_{1}^{i},x_{2}^{i},...,x_{n}^{i}]\in\mathbb{R}^{d_{i}\times n}$ & Initial attribute matrix in view $i$ \\
$\overline{X}^{i}\in\mathbb{R}^{d_{i}\times m_{i}}$ & Attribute matrix of anchors in view $i$ \\
$A^{i}\in\mathbb{R}^{n\times n}$ & Adjacency matrix in view $i$ \\
$S^{i}\in\mathbb{R}^{n\times n}$ & Attribute similarity matrix of samples in view $i$ \\
$\overline{S}^{i}\in\mathbb{R}^{n\times m_{i}}$ & Attribute similarity matrix between samples and anchors in view $i$ \\
$\tilde{S}^{i}\in\mathbb{R}^{m_{i}\times m_{i}}$ & Structural similarity matrix of anchors in view $i$ \\
$M^{i}\in\mathbb{R}^{n\times n}$ & Symmetric similarity matrix of samples in view $i$ \\
$H\in\mathbb{R}^{n\times k}$ & Similarity between samples and clusters \\
$F^{i}\in\mathbb{R}^{m_{i}\times k}$ & Similarity between anchors and clusters in view $i$ \\
$\alpha$ & Balance parameter \\
$\theta$ & Anchor proportion \\
\midrule
\end{tabular}
\caption{Main notations}
\label{notations}
\end{table}

\subsection{Multi-view subspace clustering based on anchors}
For clustering on a single view, assume the original data consists of $n$ nodes, each with a $d$-dimensional attribute vector, forming a matrix $X=[x_{1},x_{2},...,x_{n}]\in\mathbb{R}^{d\times n}$. To compute the attribute similarity between nodes, one can employ $n$ vectors as a basis to find a linear combination of them that approximates the original vectors with minimal reconstruction loss. The coefficients of this combination represent the similarity between nodes \cite{liu2012robust,elhamifar2013sparse,KANG2019510}. It is also crucial to ensure that the coefficients do not only favor the original vectors. Hence, we have

\begin{equation}
    \begin{aligned}
        \underset{S}{\text{min}} \quad & ||X-XS||_{F}^{2}+\alpha f(S) \\
        \text{s.t.} \quad & S \geq 0\ ,\ S\mathbf{1}=\mathbf{1},
    \end{aligned}
\end{equation}

where $S\in\mathbb{R}^{n\times n}$ is the similarity matrix for the $n$ nodes. To prevent $S$ from becoming an identity matrix directly, a regularization term is added, with $\alpha\geq0$ as a balance parameter, and $f(\cdot)$ as a regularization function.\par
When considering multi-view clustering, assume there are $v$ views, with the attribute matrix on each view represented as $X^{i}\in\mathbb{R}^{d_{i}\times n}$, where $d_{i}$ is the dimension of attributes for the $i$-th view \cite{gao2015multi,KANG2020105102}. The subspace clustering for this situation is described as
\begin{equation}
    \begin{aligned}
        \underset{S^{i}}{\text{min}} \quad & \sum_{i=1}^{v}||X^{i}-X^{i}S^{i}||_{F}^{2}+\alpha f(S^{i}) \\
        \text{s.t.} \quad & S^{i} \geq 0\ ,\ S^{i}\mathbf{1}=\mathbf{1},
    \end{aligned}
\end{equation}
where $S^{i}\in\mathbb{R}^{n\times n}$ is the attribute similarity matrix on the $i$-th view.\par
When reducing the complexity of multi-view clustering using the anchor method, the original attribute matrix for $n$ nodes $X^{i}\in\mathbb{R}^{d_{i}\times n}$ is transformed to $\overline{X}^{i}\in\mathbb{R}^{d_{i}\times m_{i}}$, where $m_{i}$ is the number of selected anchors $(m_{i} \ll n)$. Consequently, the subspace clustering yields a similarity matrix between $n$ original nodes and $m_{i}$ anchors, with the desired matrix being $\overline{S}^{i}\in\mathbb{R}^{n\times m_{i}}$. According to the regularization function given in \cite{kang2020large}, the problem becomes
\begin{equation}\label{3}
    \begin{aligned}
        \underset{\overline{S}^{i}}{\text{min}} \quad & \sum_{i=1}^{v}||X^{i}-\overline{X}^{i}(\overline{S}^{i})^{T})||_{F}^{2}+\alpha||\overline{S}^{i}||_{F}^{2} \\
        \text{s.t.} \quad & \overline{S}^{i} \geq 0\ ,\ (\overline{S}^{i})^{T}\mathbf{1}=\mathbf{1}.
    \end{aligned}
\end{equation}
\subsection{Multi-view spectral clustering via integrating nonnegative embedding and spectral embedding(NESE)}
Matrix decomposition is one method to solve clustering problems, with the most fundamental being the decomposition of the given initial attribute matrix, namely the Nonnegative Matrix Factorization (NMF) \cite{kim2008nonnegative}. However, this method is only suitable for cases where the underlying classification of samples is linear in the dimensional space. For nonlinear distributions of samples, we need to consider processing the similarity matrix. There are many methods to derive the similarity matrix from the initial data, generally resulting in a symmetric non-negative matrix $M\in\mathbb{R}^{n\times n}_{+}$, hence the SymNMF method \cite{kuang2012symmetric}. For a given dimension $k<n$, the goal is to find $H\in\mathbb{R}^{n\times k}_{+}$ such that
\begin{equation}
    \underset{H\geq 0}{\text{min}} \quad ||M-HH^{T}||_{F}^{2}.
\end{equation}
$H$ can be interpreted as the similarity between $n$ samples and $k$ clusters, where each node belongs to the cluster corresponding to the maximum value in its row vector in $H$.\par
Similar to the optimization problem above, the following optimization problem can also be used for clustering:
\begin{equation}
    \underset{H^{T}H=I}{\text{min}} \quad ||M-HH^{T}||_{F}^{2}.
\end{equation}
Furthermore, it can be demonstrated that, in essence, this problem can be transformed into spectral clustering \cite{hu2020multi}. If $H^{*}$ represents the spectral embedding, then applying K-means or spectral rotation to the row vectors of $H^{*}$ for discretization yields the clustering results. Combining the objectives of SymNMF and the aforementioned spectral clustering leads to
\begin{equation}
    \begin{aligned}
        \underset{F,H}{\text{min}} \quad & ||M-HF^{T}||_{F}^{2} \\
        \text{s.t.} \quad & H\geq0\ ,\ F^{T}F=I,
    \end{aligned}
\end{equation}
where $F$ represents the spectral embedding and $H$ denotes the non-negative embedding. In the context of multi-view situations, this leads to the objective function of the algorithm NESE \cite{hu2020multi}:
\begin{equation}\label{4}
    \begin{aligned}
        \underset{\left\{F^{i}\right\}_{i=1}^{v},H}{\text{min}} \quad & \sum_{i=1}^{v}||M^{i}-H(F^{i})^{T}||_{F}^{2} \\
        \text{s.t.} \quad & H\geq0\ ,\ (F^{i})^{T}F^{i}=I,
    \end{aligned}
\end{equation}
where $H$ represents the similarity between $n$ nodes and $k$ clusters, and $F^{i}$ denotes the similarity between anchors and clusters in each view.

\section{Proposed Method}
In this section, we provide a detailed introduction to the AAS method. Assume there are $n$ nodes, each having independent attribute vectors across $v$ views, constituting matrices $X^{i}=[x_{1}^{i},x_{2}^{i},...,x_{n}^{i}]\in\mathbb{R}^{d_{i}\times n}$, where $d_{i}$ is the dimension of attributes for nodes in the $i$-th view. Additionally, there exists a directed structure between the nodes, with adjacency matrix $A^{i}\in\mathbb{R}^{n\times n},i=1,...,v$.
\subsection{Structural Similarity of Anchors}
\subsubsection{Anchor selection}
\par
Firstly, the following text proposes a strategy aimed at judiciously selecting a certain number of anchors within each view, ensuring they are representative of other nodes while avoiding overly concentrated distributions in the network space to broadly cover the various clusters of the underlying data. Previous research often involves selecting high-degree nodes as anchors after aggregating all views or conducting an initial K-means step in each view with cluster centers as anchors... However, these methods typically restrict the number of anchors to be the same across all views, or even use a set of identical anchors, which to some extent limits the flexibility in choosing anchors. Below is introduced a directed structure-based method for selecting anchors. This method not only allows for inconsistency in the number of anchors across views but also considers relative balance in cluster distribution.\par
In real life, most directed networks are not strongly connected, and nodes within a strongly connected component are more likely to belong to the same cluster, as information can be exchanged among them, just as connected components in an undirected graph naturally form clusters. We first prove the following theorem regarding strongly connected components in directed graphs, with details provided in Appendix A.
\begin{theorem}
    For the directed graph $G$ with its adjacency matrix denoted by $A$ and the out-degree matrix by $D^{+}$, assume all out-degrees are nonzero. Let $P = \left(D^{+}\right)^{-1}A$ represent the probability transition matrix of a Markov process on the directed graph. The vector $\phi$ is defined as the unit left eigenvector corresponding to the spectral radius of $1$ for $P$. The non-zero elements of $\phi$ correspond to nodes within $G$ that are part of strongly connected components with only incoming edges. Furthermore, the number of linearly independent vectors $\phi$ is equal to the number of strongly connected components in $G$ that have only incoming edges.
\end{theorem}
Based on the aforementioned theorem, we can iteratively compute $\phi$ for the remaining graph, identifying its non-zero nodes' connected components as the strongly connected components of the original graph. By removing these non-zero nodes and continuing this process, we can identify all strongly connected components within the original directed graph. The specific algorithmic procedure is detailed in Appendix A. From the description of the above algorithm, it is clear that the non-zero values of each computed $\phi$ correspond to the centrality of the eigenvector for the non-zero nodes, meaning the algorithm simultaneously identifies the strongly connected components of the directed graph and the centrality of the eigenvector for nodes within each component. Therefore, it is reasonable to use the centrality of the eigenvector obtained for each strongly connected component as a criterion for selecting anchors.\par
By substituting the adjacency matrix $A^{i}$ of view $i$ into the aforementioned algorithm, all strongly connected components can be determined.
\begin{equation}
    SC^{i}=\left\{C^{i}_{1},C^{i}_{2},...,C^{i}_{c_{i}}\right\},
\end{equation}
where $c_{i}$ represents the number of strongly connected components in the $i$-th view, and $C^{i}_{j}, j=1,2,...,c_{i}$ denotes the sets of nodes in each strongly connected component of the directed graph on the $i$-th view. To ensure a uniform distribution of anchors, we select a proportion $\theta \in (0,1)$ of nodes as anchors on each strongly connected component. This proportion is consistent across all strongly connected components within a fixed view and across all views. However, due to rounding errors and the varying number of strongly connected components in each view, the number of anchors selected in each view also differs, denoted as $m_{i}$ for the $i$-th view.
\par
\subsubsection{Structural Similarity of Anchors}
On view $i$, contracting the strongly connected components $C^{i}_{1}, C^{i}_{2}, ..., C^{i}_{c_{i}}$ into single nodes results in the condensation graph, which is a directed acyclic graph with $c_{i}$ nodes. Let the matrix $a \in \mathbb{R}^{c_{i} \times c_{i}}$ represent the similarity matrix of nodes in the condensation graph, and the matrix $c \in \mathbb{R}^{m_{i} \times m_{i}}$ denote the similarity matrix of anchors in the original directed graph, where the similarity is measured by the reciprocal of the number of nodes in the shortest directed path. Assuming the similarity between two nodes is unordered, matrices $a$ and $c$ are redefined as the sum of themselves and their transposes. Matrix $b \in \mathbb{R}^{c_{i} \times m_{i}}$ records the information of the strongly connected components to which the anchors belong, specifically
\begin{equation}
    b_{jk}=\left\{
    \begin{aligned}
        1,\quad & k\in C^{i}_{j}\\
        0,\quad & k\notin C^{i}_{j}.
    \end{aligned}
    \right.
\end{equation}\par
To enhance the effect of strongly connected components in the anchor similarity matrix $c$, thereby increasing the similarity among nodes within the same component, we utilize the matrix $b$, which indicates the belonging of anchors to strongly connected components. We expand and transform the strongly connected component similarity matrix $a$ to the same dimension as $c$, and then perform element-wise multiplication with $c$, i.e., conduct the Hadamard product:
\begin{equation}
    \Tilde{S}^{i}=(b^{T}ab)\odot c.
\end{equation}
Then, $\Tilde{S}^{i} \in \mathbb{R}^{m_{i} \times m_{i}}$ becomes the structural similarity matrix for the $m_{i}$ anchors on the $i$-th view.

\subsection{Fusion Framework}
This subsection integrates the optimization problem and variables discussed above into a unified framework. For calculating the similarity between nodes and anchors on each view, we continue to use (\ref{3}). Next, it is necessary to merge the attribute and structural similarities for each view. $\overline{S}^{i}$ represents the attribute similarity matrix between $n$ samples and $m_{i}$ anchors on the $i$-th view, which drags the nodes to be classified closer to the anchors by increasing their values based on attributes; $\Tilde{S}^{i}$ is the structural similarity matrix for $m_{i}$ anchors on the $i$-th view, which draws anchors of the same cluster closer to each other based on structural information. Thus, $\overline{S}^{i}\Tilde{S}^{i} \in \mathbb{R}^{n \times m_{i}}$ is the similarity matrix between samples and anchors that integrates attribute and structural information, including the underlying classification information reflected by strongly connected components. Substituting the matrix $\overline{S}^{i}\Tilde{S}^{i}$ into the NESE clustering mentioned in Section 2 and considering the different weights of each view, transforms problem (\ref{4}) into
\begin{equation}
    \begin{aligned}
        \underset{\left\{F^{i}\right\}_{i=1}^{v},H,p_{i}}{\text{min}} \quad & \sum_{i=1}^{v}\frac{1}{p_{i}}||\overline{S}^{i}\Tilde{S}^{i}-H(F^{i})^{T}||_{F}^{2} \\
        \text{s.t.} \quad & H\geq0\ ,\ (F^{i})^{T}F^{i}=I\ ,\ p_{i}\geq0\ ,\ \sum_{i=1}^{v}p_{i}=1.
    \end{aligned}
\end{equation}
To avoid suboptimal solutions resulting from solving optimization problems in separate steps, the calculation of the attribute similarity matrix and clustering steps are integrated into a unified optimization framework, specifically
\begin{equation}\label{14}
    \begin{aligned}
        \underset{\overline{S}^{i},H,F^{i},p_{i}}{\text{min}} \quad & \sum_{i=1}^{v}||X^{i}-\overline{X}^{i}(\overline{S}^{i})^{T})||_{F}^{2}+\alpha||\overline{S}^{i}||_{F}^{2}+\sum_{i=1}^{v}\frac{1}{p_{i}}||\overline{S}^{i}\Tilde{S}^{i}-H(F^{i})^{T}||_{F}^{2} \\
        \text{s.t.} \quad & \overline{S}^{i} \geq 0,(\overline{S}^{i})^{T}\mathbf{1}=\mathbf{1},H\geq0,(F^{i})^{T}F^{i}=I,p_{i}\geq0,\sum_{i=1}^{v}p_{i}=1.
    \end{aligned}
\end{equation}

It should be emphasized that Equation (12) introduces $\Tilde{S}^{i}$ to redefine the node similarity matrix of the original NESE method, expanding its symmetric similarity matrix $M$ for all nodes to an asymmetric similarity matrix $\overline{S}^{i}\Tilde{S}^{i}$ for all nodes and anchors, with its effectiveness validated in subsequent sections. Additionally, self-learning weights $\frac{1}{p_{i}}$ are set for each view to highlight the differences in the impact of various views on the clustering results.

\section{Optimization of proposed method}
This section presents the solution algorithm for the optimization problem (\ref{14}), and analyzes its algorithmic time complexity and convergence.
\subsection{Optimization Algorithm}
Solving the optimization problem (\ref{14}) is not straightforward, as both the control variables and the objective function are intertwined, making it unsuitable for direct optimization algorithms. Here, we employ an alternating iterative method. Moreover, not all control variables are needed for our purposes, so the solution process focuses only on the variables that are required.\par
\subsubsection{Computation of $\overline{S}^{i}$}
For fixed $H, F^{i}, p_{i}$, by combining the preceding and succeeding sum terms, solving problem (\ref{14}) entails solving the following for any $i$:
\begin{equation}\label{13}
    \begin{aligned}
        \underset{\overline{S}^{i}}{\text{min}} \quad & ||X^{i}-\overline{X}^{i}(\overline{S}^{i})^{T})||_{F}^{2}+\alpha||\overline{S}^{i}||_{F}^{2}+\frac{1}{p_{i}}||\overline{S}^{i}\Tilde{S}^{i}-H(F^{i})^{T}||_{F}^{2} \\
        \text{s.t.} \quad & \overline{S}^{i} \geq 0,(\overline{S}^{i})^{T}\mathbf{1}=\mathbf{1}.
    \end{aligned}
\end{equation}
The objective function of the aforementioned problem, written in quadratic form, is equivalent to
\begin{equation}
    Trace\ [\ \frac{1}{2}\overline{S}^{i}A(\overline{S}^{i})^{T}+f^{T}(\overline{S}^{i})^{T}+g((\overline{S}^{i})^{T})\ ]
\end{equation}
where $A = 2(\overline{X}^{i})^{T}\overline{X}^{i} + 2\alpha I_{m_{i}} + 2\frac{1}{p_{i}}\Tilde{S}^{i}(\Tilde{S}^{i})^{T}$, and $f = -2(\overline{X}^{i})^{T}X^{i} - 2\frac{1}{p_{i}}\Tilde{S}^{i}F^{i}H^{T}$. $g((\overline{S}^{i})^{T})$ is a term that is independent of $(\overline{X}^{i})^{T}$. Thus, considering $(\overline{S}^{i})^{T}$ as the control variable, solving problem (\ref{13}) is equivalent to solving the following quadratic optimization problem
\begin{equation}\label{15}
    \begin{aligned}
        \underset{(\overline{S}^{i})^{T}}{\text{min}} \quad & \frac{1}{2}\overline{S}^{i}A(\overline{S}^{i})^{T}+f^{T}(\overline{S}^{i})^{T} \\
        \text{s.t.} \quad & \mathbf{0}\leq (\overline{S}^{i})^{T} \leq \mathbf{1},(\overline{S}^{i})^{T}\mathbf{1}=\mathbf{1}.
    \end{aligned}
\end{equation}
where $\mathbf{1}$ is an all-ones column vector, with its
 dimension adapted to the context as required.
\subsubsection{Computation of H}
When $\overline{S}^{i}$ is fixed, the optimization problem (\ref{14}) becomes
\begin{equation}\label{16}
    \begin{aligned}
        \underset{\left\{F^{i}\right\}_{i=1}^{v},H,p_{i}}{\text{min}} \quad & \sum_{i=1}^{v}\frac{1}{p_{i}}||\overline{S}^{i}\Tilde{S}^{i}-H(F^{i})^{T}||_{F}^{2} \\
        \text{s.t.} \quad & H\geq0\ ,\ (F^{i})^{T}F^{i}=I\ ,\ p_{i}\geq0\ ,\ \sum_{i=1}^{v}p_{i}=1.
    \end{aligned}
\end{equation}
For solving $H$, it is proven in Appendix B that the aforementioned optimization problem is equivalent to the following two optimization problems:
\begin{equation}\label{17}
    \begin{aligned}
        \underset{\left\{F^{i}\right\}_{i=1}^{v},H}{\text{min}} \quad & \sum_{i=1}^{v}||\overline{S}^{i}\Tilde{S}^{i}-H(F^{i})^{T}||_{F} \\
        \text{s.t.} \quad & H\geq0\ ,\ (F^{i})^{T}F^{i}=I\ .
    \end{aligned}
\end{equation}
and
\begin{equation}\label{18}
    \begin{aligned}
        \underset{\left\{F^{i}\right\}_{i=1}^{v},H}{\text{min}} \quad & \sum_{i=1}^{v}\frac{1}{p_{i}}||\overline{S}^{i}\Tilde{S}^{i}-H(F^{i})^{T}||_{F}^{2} \\
        \text{s.t.} \quad & H\geq0\ ,\ (F^{i})^{T}F^{i}=I\ ,\ p_{i}=||\overline{S}^{i}\Tilde{S}^{i}-H_{*}(F_{*}^{i})^{T}||_{F}.
    \end{aligned}
\end{equation}
where $H_{*}$ and $F_{*}^{i}$ are the optimal solutions for $H$ and $F^{i}$, respectively. The optimization problem (\ref{18}) can be solved using the inexact Majorization-Minimization (MM) method, with the specific process detailed in \cite{hu2020multi}.

The proposed optimization algorithm for solving Eqs. (\ref{14}) is summarized below.

\begin{algorithm}
	\renewcommand{\algorithmicrequire}{\textbf{Input:}}
	\renewcommand{\algorithmicensure}{\textbf{Output:}}
	\caption{Optimization Algorithm of AAS}
	\label{alg1}
	\begin{algorithmic}[1]
        \REQUIRE Attribute matrices \(X^1, X^2, \dots, X^v\) for \(v\) views; structure similarity matrices \(\Tilde{S}^{1}, \Tilde{S}^{2}, \dots, \Tilde{S}^{v}\) for \(v\) views; number of clusters \(k\); parameters \(\alpha > 0\).
        \ENSURE  \(\overline{S}^{i}, H, F^{i}, p_{i}\).
		\STATE  Initialize reciprocal of weight for each view, \(p_{i} = \frac{1}{v}\),
		\STATE  Initialize the clustering results \(H\) randomly,
        \STATE  Initialize attribute similarity matrices \(\overline{S}^{i}\) so that each of its entries is \( \frac{1}{n} \),
        \STATE  Initialize \(F^{i}\) by the spectral embedding of \( (\overline{S}^{i}\Tilde{S}^{i})^{\top} \overline{S}^{i}\Tilde{S}^{i} \).
		\REPEAT
		\STATE Update \(\overline{S}^{i}\) solving (\ref{15}) via SQP,
		\REPEAT
		\STATE Update \(H, F^{i}, p_{i}\) solving (\ref{18}) via inexact-MM,
		\UNTIL \( t > t_{\max} \)
		\UNTIL stopping criterion is met.
	\end{algorithmic}
\end{algorithm}
\subsection{Computation complexity and convergence analysis}
The time complexity of solving Algorithm 1 is divided into the initialization phase and the iterative optimization phase. Let $m=\max_{i}\left\{m_{i}\right\}$, and $m \ll n$. In the initialization phase, the primary computational cost is $\mathcal{O}(vn^{2})$ for $\overline{S}^{i}$.For the alternating iterative optimization phase, solving the attribute similarity between nodes and anchors using SQP incurs a cost of $\mathcal{O}(nm^{3}v)$. After obtaining a new $\overline{S}^{i}$multiplying it by the structural similarity matrix $\Tilde{S}^{i}$ costs $\mathcal{O}(nm^{2}v)$. Solving the optimization problem (\ref{18}) to obtain \(H, F^{i}, p_{i}\) costs $\mathcal{O}(nk^{2})$,and this step needs to be repeated $t_{max}$ times. Assuming the outer loop requires $t$ iterations to reach the stopping condition, the time complexity of the iterative part of the algorithm is:

\begin{equation*}
    \mathcal{O}(((nm^{2}v+nk^{2})t_{max}+nm^{3}v)t).
\end{equation*}

Note that $m \ll n$, $k \ll n$, $v \ll n$, $t_{max} \ll n$, and $t \ll n$. Thus, it can be concluded that when the data scale is very large, the primary time expenditure of the AAS algorithm is in the initialization phase. Even though the structural similarity matrix needs to be repeatedly integrated during the iterations, it does not significantly impact the overall time complexity. Additionally, for data with a large number of views to be classified, the calculations in equation (\ref{15}) can be performed in parallel across the views to save time.\par
In solving the optimization problem (\ref{14}), problems (\ref{15}) and (\ref{18}) are alternately solved.Subproblem (\ref{15}) is clearly a quadratic programming problem, and its Hessian matrix is evidently positive definite, making this subproblem convex.Problem (\ref{18}) has been proven to converge by \cite{hu2020multi}. Therefore, problem (\ref{14}) is convergent.

\section{Experiments}
In this section, we will test the proposed algorithm AAS on synthetic data to demonstrate its performance. The proposed algorithm was implemented using Python and MATLAB software environments on a system running Windows 11 with an AMD Ryzen 7 6800H with Radeon Graphics CPU @ 3.20 GHz and 16 GB RAM.

\subsection{Datasets}

In the process of data mining and information extraction from certain samples, it is common and easy to obtain both attribute information and structural information. These two types of information are helpful for the clustering analysis of samples. However, it is not common in the study of clustering algorithms to consider both simultaneously. Therefore, it is challenging to find experimental data that fits the scenario of the algorithm presented in this paper. We have only found a real-world network dataset that contains only directed structural information without any attribute information, namely "Seventh graders". Additionally, we generated network data that conforms to the AAS algorithm. The generation method is as follows:

\subsubsection{Synthetic datasets}

This paper adopts the Attribute SBM (Stochastic Block Model) \cite{stanley2019stochastic} to generate networks. The Attribute SBM model is designed for single-view networks, setting the undirected structure of the network and the attribute vector on each node according to pre-defined node clusters. When applied to the AAS algorithm, it only needs a simple generalization: setting multiple views and directed structures.

We aim to generate two networks with $v=3$ views, each containing $n=50$ nodes and $n=5000$ nodes, respectively, with both networks consisting of $k=4$ clusters. These two networks are named "Attribute SBM\_50" and "Attribute SBM\_5000" in this paper. Let $z$ be an $n$-dimensional vector representing the clusters to which the nodes belong, marked with 0, 1, 2, and 3, with the number of nodes in each cluster being 10, 15, 12, 13 and 1000, 1500, 1200, 1300, respectively. First, we outline the construction of directed edges for the network. To distinguish between different views, it is only necessary to change the probability distribution of the edges under the same cluster settings, i.e., essentially the same but with different manifestations. According to this criterion, the directed edges of the three views are set to satisfy the following distribution, noting that the indices of the adjacency matrix A are ordered, even though $A_{ij}$ and $A_{ji}$ follow the same distribution.

\begin{equation*}
    A_{ij}^{1} \sim \left\{
    \begin{aligned}
        Bernoulli(.20),\quad & if\ z_{i}=z_{j}\\
        Bernoulli(.01),\quad & if\ z_{i}\neq z_{j}
    \end{aligned}
    \right.
\end{equation*}
\begin{equation*}
    A_{ij}^{2} \sim \left\{
    \begin{aligned}
        Bernoulli(.20),\quad & if\ z_{i}=z_{j}\\
        Bernoulli(.03),\quad & if\ z_{i}\neq z_{j}
    \end{aligned}
    \right.
\end{equation*}
\begin{equation*}
    A_{ij}^{3} \sim \left\{
    \begin{aligned}
        Bernoulli(.25),\quad & if\ z_{i}=z_{j}\\
        Bernoulli(.03),\quad & if\ z_{i}\neq z_{j}
    \end{aligned}
    \right.
\end{equation*}

\begin{figure}[H]
  \centering
    \includegraphics[width=0.7\linewidth]{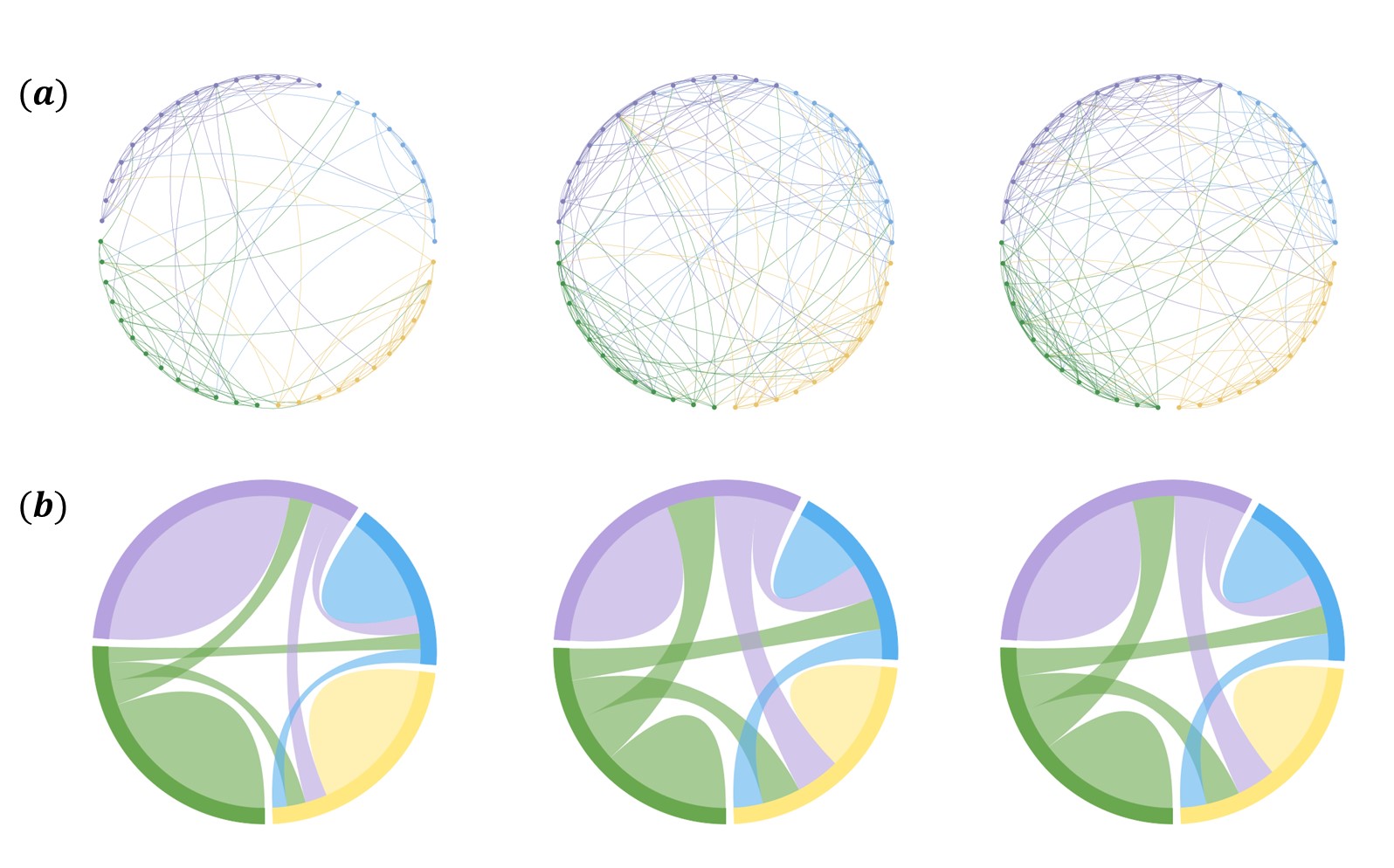}
  \caption{Network visualization of 3 views on Attribute SBM\_50 and Attribute SBM\_5000}
  \label{network1}
\end{figure}

To clearly observe the differences between the intra-cluster edges and inter-cluster edges in the generated network, we use the following graph to illustrate the structural information. Fig.\ref{network1} presents the visualization of the constructed network structure. (a) displays three views of the "Attribute SBM\_50", while (b) shows three views of the "Attribute SBM\_5000". 
Using a chord diagram layout makes the graph structure more intuitive. For dataset "Attribute SBM\_5000", a more illustrative chord diagram is adopted due to the large number of nodes and edges. 

For randomly generated directed network structures or real-world directed network data, it is necessary to preprocess the data in cases of noise or excessive sparsity. Noise nodes can be removed by evaluating the local clustering coefficient \cite{li2017clustering}, discarding nodes with values that are too low. Set reasonable edge weights and use methods such as threshold filtering to remove abnormal edges. When the directed network structure is relatively sparse, node similarity or link prediction methods can be used to predict and complete missing edges. After the above processing, the directed network data becomes more credible, and the accuracy and reliability of the clustering results are enhanced.

\begin{figure}[H]
  \centering
    \includegraphics[width=\linewidth]{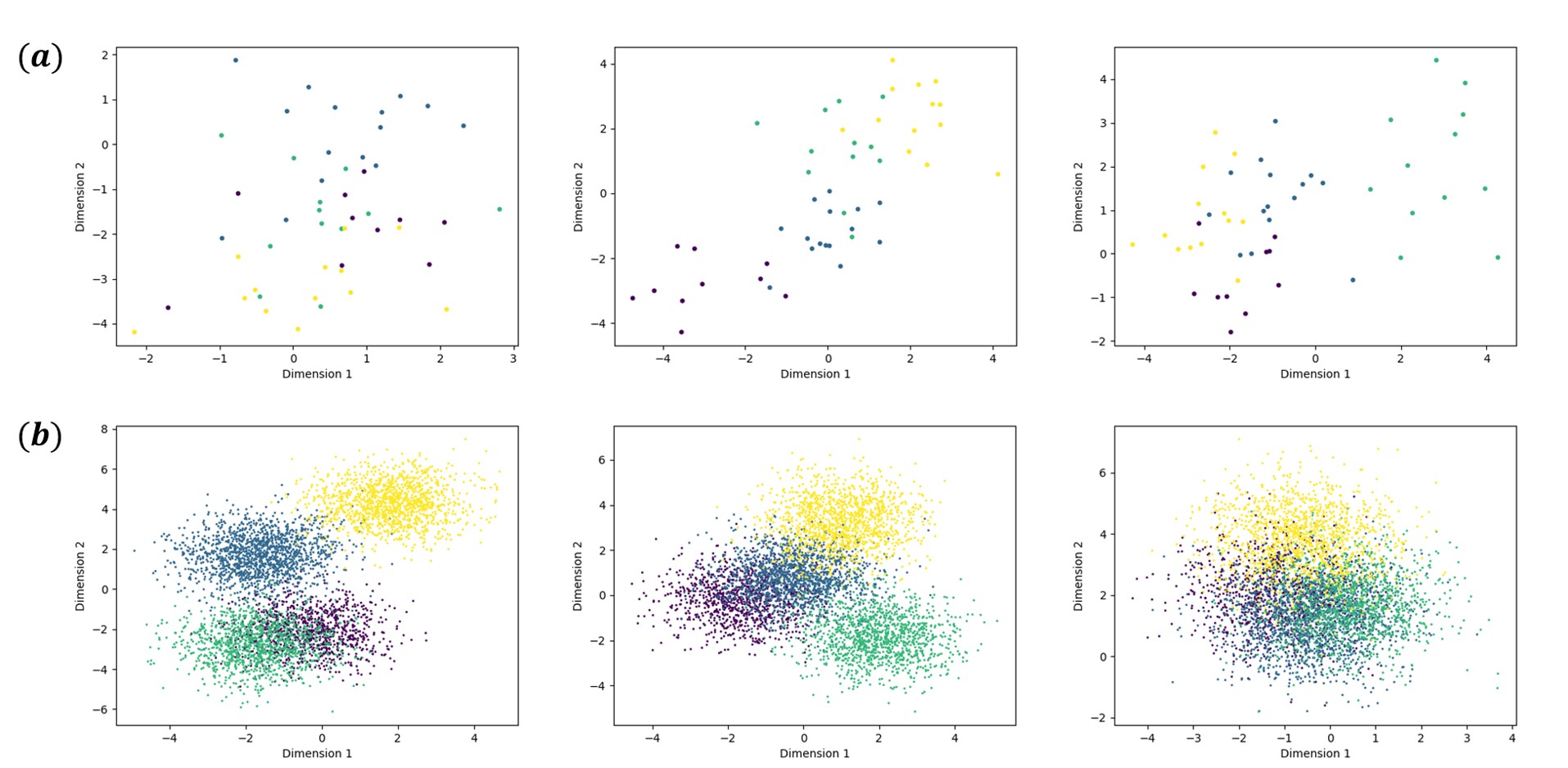}
  \caption{Attributes visualization of 3 views on Attribute SBM\_50 and Attribute SBM\_5000}
  \label{attribute}
\end{figure}

For the attribute vectors in the three views, we will assume for visualization purposes that the attributes in all three views are two-dimensional. The following description is applicable to all views. Firstly, we assign a mean vector $\mu_{k}$ to each cluster, where each element follows a Gaussian distribution $\mathcal{N}(0,2)$. Secondly, we set the covariance matrix $\Sigma_{k}$ for each cluster as a diagonal matrix $diag(1.25,1.25)$. Thus, the attribute vector of nodes within each cluster follows a multivariate Gaussian distribution $\mathcal{N}(\mu_{k},\Sigma_{k})$. Based on the generated attribute vectors, the nodes on the three views are plotted on the scatter plot in Fig.\ref{attribute}, where nodes of the same color belong to the same cluster. It can be observed that nodes of the same cluster are relatively closer to each other, but it is difficult to determine the clustering boundaries based solely on the attribute nodes.

\subsubsection{Real-world datasets}

"Seventh graders" dataset comes from a school in Victoria, Australia, and includes 29 students. Each student was asked to answer three questions: Who are your best friends, who do you get on well with, and who do you like to work with? The answers to these three questions formed three directed graphs among the 29 students, which we treat as three views. Additional information provided includes each student’s gender, which naturally divides them into two underlying clusters. In summary, "Seventh graders" dataset contains only directed structural information across three views without any attribute information. The dataset will be used in the ablation study to verify the role of structural information in the AAS algorithm.

\subsection{Baseline algorithms and metrics}
There are hardly any algorithms in the current research domain that address the problem setting identical to that discussed in this paper. The majority of existing research on multi-view clustering primarily focuses on independently utilizing either the attribute information or the structural information of the samples, rather than a fusion of the two. Hence, the data generated above can only be partially applied in these studies, without fully utilizing all available information. To demonstrate the performance of the proposed AAS algorithm when sample information is fully provided, we compared it with existing algorithms. As part of the comparison, we first attempted to apply the most basic clustering method  K-means algorithm  on each view and selected the best-performing results from multiple views. Additionally, we selected several of the latest open-source multi-view clustering algorithms for comparison, including NESE \cite{hu2020multi}, GMC \cite{wang2019gmc}, LMVSC \cite{kang2020large}, SMC \cite{liu2022scalable}, OMSC \cite{chen2022efficient} and CAMVC \cite{zhang2024learning}.

To comprehensively evaluate the clustering performance of the aforementioned algorithms, we adopted three widely recognized clustering performance metrics \cite{zhan2018multiview}: ACC (Clustering Accuracy), NMI (Normalized Mutual Information), and Purity. These metrics are extensively used in the research of clustering algorithms because they offer a comprehensive assessment of the quality of clustering results from different perspectives. ACC primarily measures the consistency between the clustering results and the true labels; NMI assesses the degree of information shared between the clustering results and the true distribution; while Purity focuses on the purity of the clusters post-clustering, i.e., the proportion of the majority true label within each cluster. Through the comprehensive assessment provided by these metrics, we can gain a more accurate understanding of the performance of each algorithm within the experimental environment.

\subsection{Performance evaluation}

\begin{figure}[H]
  \centering
    \includegraphics[width=\linewidth]{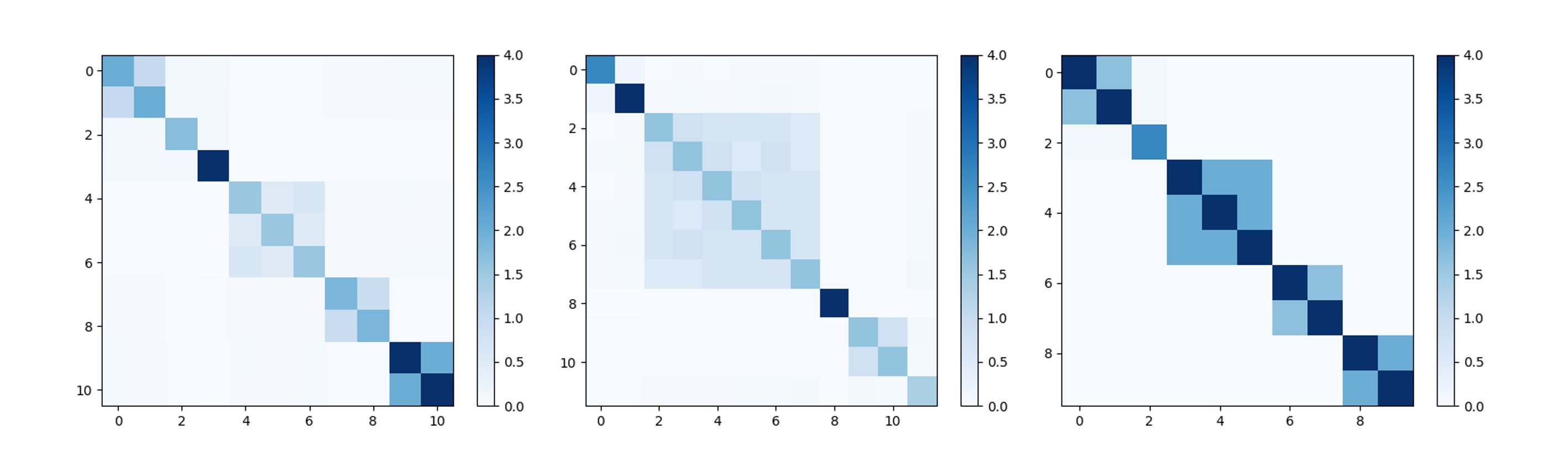}
  \caption{Visualization of anchor similarity matrices on 3 views}
  \label{anchor}
\end{figure}

We first validate the anchor structure similarity matrix defined earlier based on the synthetic dataset to ascertain its rationality. Fig.\ref{anchor} takes "Attribute SBM\_50" as an example. The matrices illustrate the similarity of anchor structures derived from their respective directed network structures. A darker color indicates higher similarity. Besides the diagonal, which shows each anchor is most similar to itself. It is observable that the first graph contains six diagonal blocks. Each block represents a strongly connected component within the original directed network, indicating that the directed network structure of the first view comprises six strongly connected components. The second graph displays a large diagonal block, a diagonal block with only two elements, and four singular element blocks; the third graph features five diagonal blocks. If matrix coordinates are within a diagonal block, the corresponding anchors reside within the same strongly connected component. Adhering to the underlying premise of this study, nodes within the same strongly connected component tend to belong to the same cluster. It follows that corresponding anchors are inclined to belong to the same cluster. In generating data, we arranged node numbers by cluster, meaning nodes within the same cluster have proximate numbers. This suggests that nodes within the same strongly connected component have close numbers, thus forming diagonal blocks in the figure below, while elements far from the diagonal have low similarity. This is highly consistent with the requirements of the AAS algorithm established in this paper.

\begin{table}[H]
\centering
\begin{tabular}{ccccc}
\toprule
Dataset & Algorithm & ACC & NMI & Purity \\
\midrule
Attribute SBM\_50 & K-means & 0.763(0.02) & 0.651(0.01) & 0.763(0.02) \\
 & NESE & 0.865($\rightarrow$0) & 0.764($\rightarrow$0) & 0.865($\rightarrow$0) \\
 & GMC & 0.857($\rightarrow$0) & 0.752($\rightarrow$0) & 0.857($\rightarrow$0) \\
 & LMVSC & 0.812($\rightarrow$0) & 0.739($\rightarrow$0) & 0.812($\rightarrow$0) \\
 & SMC & 0.848($\rightarrow$0) & 0.782($\rightarrow$0) & 0.853($\rightarrow$0) \\
 & OMSC & 0.889($\rightarrow$0) & 0.805($\rightarrow$0) & 0.892($\rightarrow$0) \\
 & CAMVC & 0.897($\rightarrow$0) & 0.804($\rightarrow$0) & 0.904($\rightarrow$0) \\
 & AAS & \textbf{0.932(0.01)} & \textbf{0.852(0.03)} & \textbf{0.932(0.01)} \\
 Attribute SBM\_5000 & K-means & 0.594(0.03) & 0.543(0.01) & 0.594(0.03) \\
 & NESE & 0.808($\rightarrow$0) & 0.768($\rightarrow$0) & 0.811($\rightarrow$0) \\
 & GMC & 0.786($\rightarrow$0) & 0.744($\rightarrow$0) & 0.787($\rightarrow$0) \\
 & LMVSC & 0.684($\rightarrow$0) & 0.656($\rightarrow$0) & 0.693($\rightarrow$0) \\
 & SMC & 0.719($\rightarrow$0) & 0.680($\rightarrow$0) & 0.719($\rightarrow$0) \\
 & OMSC & 0.762($\rightarrow$0) & 0.727($\rightarrow$0) & 0.762($\rightarrow$0) \\
 & CAMVC & 0.810($\rightarrow$0) & 0.771($\rightarrow$0) & 0.810($\rightarrow$0) \\
 & AAS & \textbf{0.841(0.02)} & \textbf{0.809(0.03)} & \textbf{0.841(0.02)} \\
\midrule
\end{tabular}
\caption{Clustering results}
\label{result}
\end{table}
The Table \ref{result} presents the clustering results on the synthetic data from Section 5.1 using the AAS algorithm proposed in this paper, along with seven other algorithms.The performance is measured using three metrics: ACC, NMI, and Purity, all ranging between 0 and 1, with higher values indicating better performance. The synthetic data were used to run each algorithm 20 times, and the results in the table represent the mean values, with standard deviations in parentheses. The maximum mean value for each metric is highlighted in bold. Observing the numerical results in the table, the following analyses can be made:

\begin{itemize}
    \item Across all three metrics, the proposed AAS algorithm exhibits the best clustering performance among all compared algorithms, with a significant improvement. However, its standard deviation is relatively large.
    \item The AAS algorithm proposed in this paper references the NESE algorithm in the step of clustering based on the similarity matrix. Compared to NESE, AAS shows a significant improvement, indicating that the method of integrating structural information to compute the similarity matrix in AAS has a positive impact on clustering results.
    \item The K-means algorithm applied in this paper selects the best results from three views; however, its essence remains single-view clustering, unlike other algorithms which perform multi-view clustering. From the numerical results, K-means shows considerable deviation from other algorithms. Apart from inherent algorithmic differences, this deviation is likely due to its lack of integration of multi-view information, utilizing only information from a single perspective.
\end{itemize}

\subsection{Computation analysis}
\subsubsection{Time comparison}
The bar chart in Fig.\ref{fig6}(a) illustrates the execution times (in seconds) of various methods on "Attribute SBM\_50", excluding the single-view K-means algorithm. The numerical values on the bars represent the time taken by each method for a single run. AAS and all baseline methods have linear time complexity, making them scalable across all datasets and applicable to large-scale networks. Our method significantly improves accuracy while exceeding some baseline methods in terms of time. 

\subsubsection{Convergence study}
\begin{figure}[htbp]
    \centering
    \begin{subfigure}[t]{0.48\textwidth}
        \centering
        \includegraphics[width=1\linewidth]{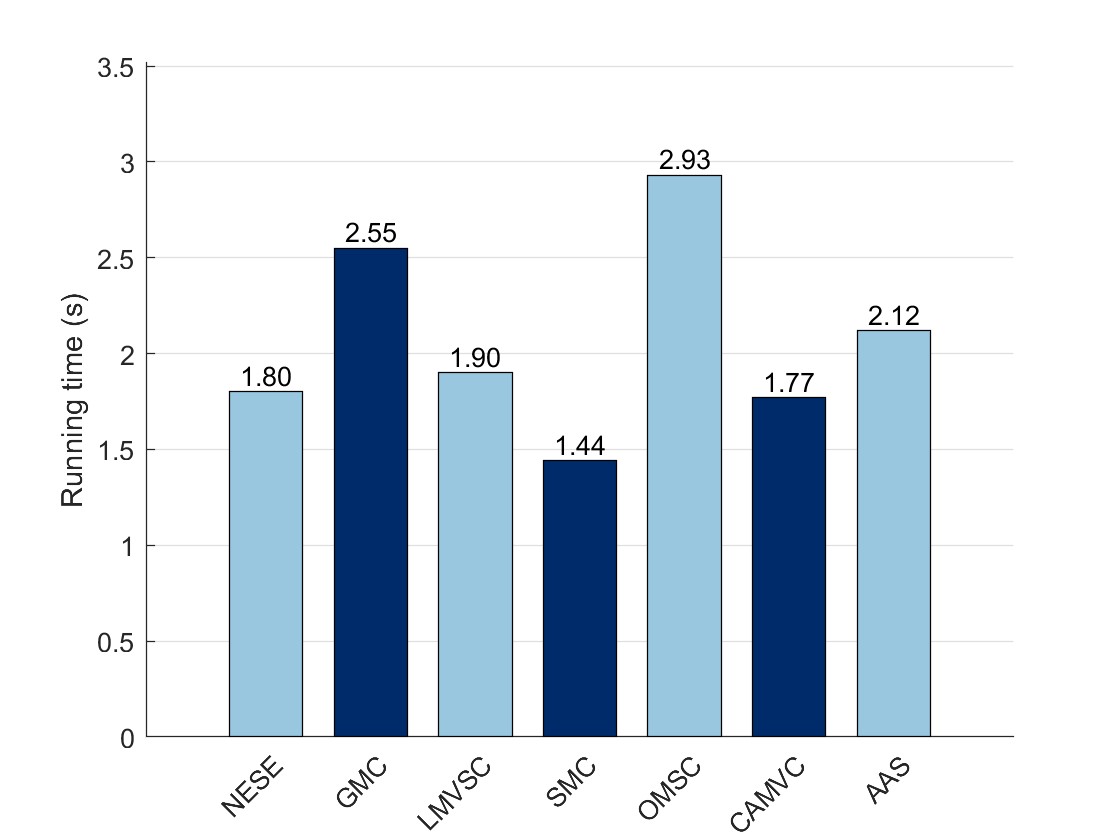}
        \captionsetup{labelformat=empty}
        \caption{(a)}
    \end{subfigure}
    \hfill
    \begin{subfigure}[t]{0.48\textwidth}
        \centering
        \includegraphics[width=1\linewidth]{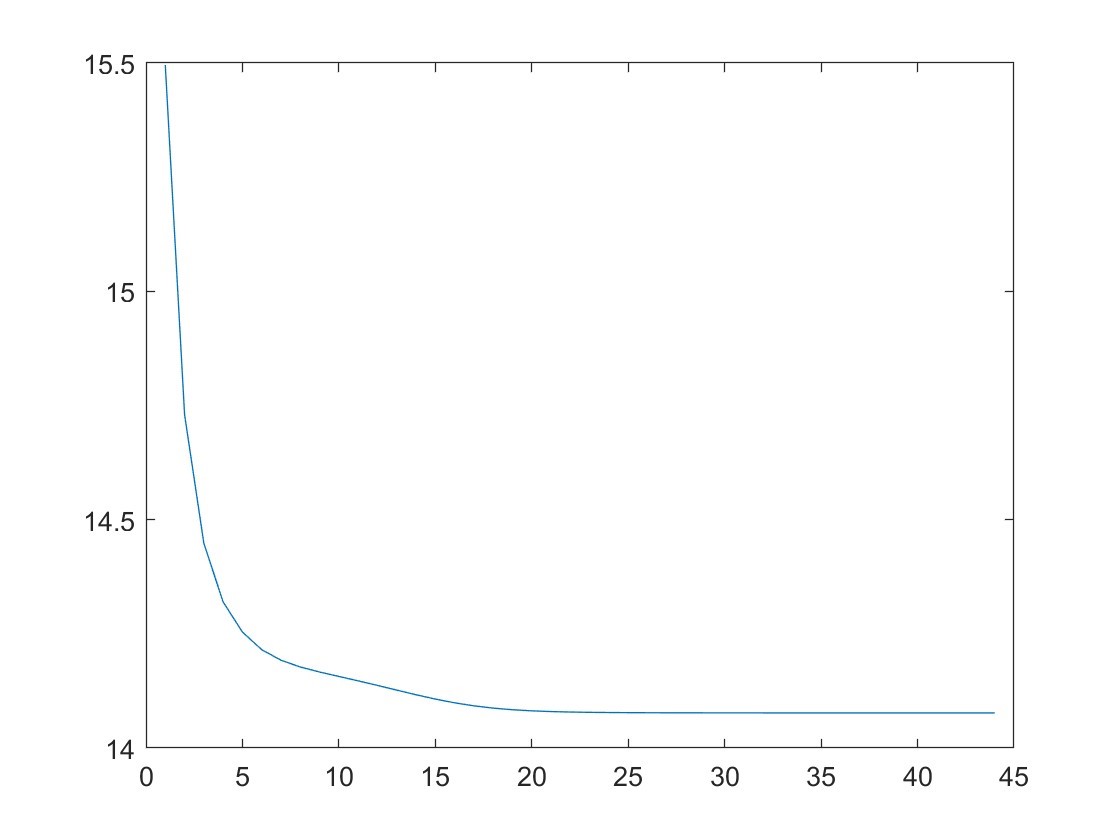}
        \captionsetup{labelformat=empty}
        \caption{(b)}
    \end{subfigure}
    \caption{Running time and convergence curve of AAS}
    \label{fig6}
\end{figure}

Fig.\ref{fig6}(b)illustrates the convergence speed of the AAS algorithm on "Attribute SBM\_50", although the vertical axis does not depict the variation of the objective function value of problem (\ref{14}). As explained in Section 4 of this paper regarding optimization algorithms, it is more convenient to track the changes in the objective function minimum value of problem (\ref{15}), which does not affect the understanding of the convergence trend and speed. From the figure, it can be observed that the objective function value of AAS decreases as the number of iterations increases, and it essentially converges within 20 iterations, indicating a fast convergence speed of AAS.

\subsubsection{Ablation study}
The AAS algorithm, compared to common multi-view clustering algorithms, includes two main highlights: the integration of directed structural information and the anchor selection based on directed structures. To demonstrate the effectiveness of these ideas for multi-view clustering outcomes, we first designed ablation studies comparing the results of the AAS algorithm with the following two models on both datasets "Attribute SBM\_50" and "Attribute SBM\_5000":\par

\textbf{Experiment AA$\_$S (removing structural information):} To explore the impact of directed structural information on the AAS algorithm, we set the anchor structural similarity matrix to an identity matrix, thus disregarding structural information and considering only attribute information like other existing algorithms. This version is referred to as AA$\_$S and compared with AAS. The results in Table \ref{ablation} show that the anchor structural similarity matrix defined in AAS significantly enhances clustering accuracy. Furthermore, since it uses the same dataset as the comparative experiments, AA$\_$S performs better relative to most other multi-view clustering algorithms that only consider attribute information, indicating that the AAS algorithm's model for clustering based on attribute information is also highly effective.

\textbf{Experiment A$\_$AS (random selection of anchors):} We propose an anchor selection strategy based on directed structural information aimed at ensuring representativeness and even distribution across the network. Many existing multi-view clustering algorithms that use anchors typically select a fixed number of anchors randomly from each view. We compared AAS with this random anchor method, A$\_$AS. The results in Table \ref{ablation} show that the anchor selection strategy has a limited impact on clustering improvement and occasionally performs worse than random selection. However, we still believe that selecting anchors in the AAS is necessary. It improves clustering performance in most cases with low computational cost and establishes a basis for defining the anchor structural similarity matrix.

\begin{table}[H]
\centering
\begin{tabular}{ccccc}
\toprule
Dataset & Algorithm & ACC & NMI & Purity \\ 
\midrule
Attribute SBM\_50 & AAS & \textbf{0.932} & \textbf{0.852} & 0.932 \\ 
 & AA$\_$S & 0.897 & 0.795 & 0.897 \\
 & A$\_$AS & 0.931 & 0.848 & \textbf{0.935} \\
 Attribute SBM\_5000 & AAS & \textbf{0.841} & \textbf{0.809} & \textbf{0.841} \\ 
 & AA$\_$S & 0.812 & 0.770 & 0.812 \\
 & A$\_$AS & 0.833 & 0.792 & 0.836 \\
\midrule
\end{tabular}
\caption{Ablation study of synthetic datasets}
\label{ablation}
\end{table}

The "Seventh graders" is a real-world dataset which contains network structures with three views but no node attributes. It's suitable for highlighting the role of the AAS algorithm in structural clustering, which is one of the main objectives of this paper. We set the node attributes of this dataset to be a vector of ones. The results of both K-means and AAS algorithms are shown in Table \ref{Seventh graders}, where K-means selects the best result among all views. All student nodes are divided into two classes, with K-means assigning all but one node to the first class. This is because the attribute information being all ones clearly does not provide any useful learning, while AAS can leverage the topological structure of the data to achieve better clustering results.

\begin{table}[H]
\centering
\begin{tabular}{cccc}
\toprule
Algorithm & ACC & NMI & Purity \\ 
\midrule
AAS & \textbf{0.882} & \textbf{0.693} & \textbf{0.891} \\ 
K-means & 0.655 & 0.052 & 0.655 \\
\midrule
\end{tabular}
\caption{Clustering results on ”Seventh graders”}
\label{Seventh graders}
\end{table}

\subsubsection{Parameter analysis}
We designed a parameter sensitivity analysis experiment on the "Attribute SBM\_50" dataset to evaluate the impact of different parameter settings on the performance of the AAS algorithm. Two main parameters were considered: the balance parameter of the regularization term in the attribute similarity matrix solution process, denoted as $\alpha$, and the proportion of anchors in each strongly connected component, denoted as $\theta$. The value ranges for these two parameters were set as follows: $\alpha \in \left\{0.001,0.01,0.1,1,10\right\}$ and $\theta \in \left\{0.1,0.2,0.3,0.4,0.5\right\}$. The AAS algorithm was run with each parameter combination, and the ACC, NMI, and Purity metrics were recorded. The experimental results are presented as bar charts below. To facilitate easier comparison of subtle differences between the results, the z-axis in the charts is set from 0.5 to 1. From the results, it can be observed that the balance parameter $\alpha$ should be controlled below 1, with 0.1 being a preferable choice. When $\alpha=10$, there is a significant decrease in all three metrics. The AAS algorithm shows relatively low sensitivity to the parameter $\theta$, the anchor ratio. However, from the perspective of algorithm complexity, $\theta$ should ideally not exceed 0.3, with 0.3 slightly outperforming 0.1 and 0.2. The above analysis indicates that the parameter values $\alpha=0.1$ and $\theta=0.3$ are the optimal settings for the current dataset and experimental conditions. Unless otherwise specified, this parameter setting is assumed throughout this paper. Through parameter sensitivity analysis, a better understanding of how each parameter affects the clustering performance of the AAS algorithm can be achieved, aiding in optimizing the algorithm's performance.

\begin{figure}[H]
    \centering
    \includegraphics[width=1\linewidth]{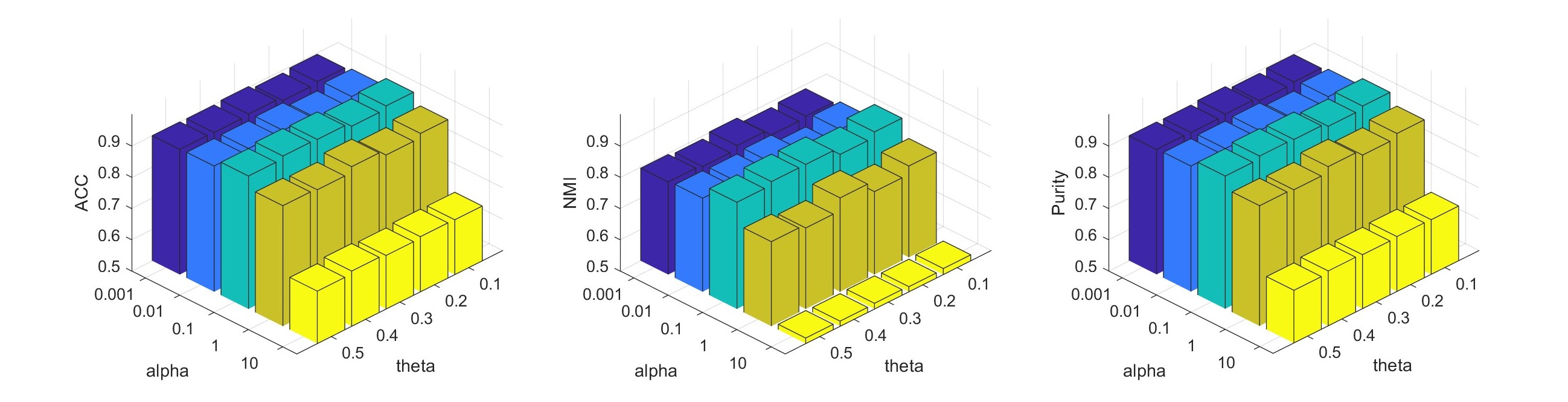}
    \caption{Parameter analysis}
\end{figure}

~\\

\section{Conclusion}
This paper introduces a new multi-view clustering algorithm, AAS, which integrates multi-views, node attributes, and directed structural information to highlight underlying clusters in the node similarity matrix. AAS operates within a comprehensive optimization framework, directly outputting a cluster result matrix without requiring post-processing steps like discretization. To address the common shortfall of structural information in multi-view clustering data, this study adapates the Attribute SBM model to suit AAS's needs. Comparative analysis with seven other methods demonstrates that effective structural integration improves clustering accuracy.

The study also recognizes limitations of the AAS algorithm. Its reliance on specific directed network structures and the increased computational cost due to the integration of structural information suggest areas for future improvement. The use of anchors as intermediaries illustrates a potential stepwise approach for enhancing similarity among clusters in different clustering contexts. With advancements in data collection and storage, the potential for using more complex datasets in classification grows, prompting further exploration into the integration and mining of varied data types.

~\\
\textbf{CRediT authorship contribution statement}\par
\ \par
\textbf{Xuetong Li:} Conceptualization, Methodology, Software, Writing-original draft. \textbf{Xiao-Dong Zhang:} Methodology, Project administration, Supervision, Writing
-review $\&$ editing.

~\\
\textbf{Declaration of competing interest}\par
\ \par
The authors declare that they have no known competing financial interests or personal relationships that could have appeared to influence the work reported in this paper.

~\\
\textbf{Data availability}\par
\ \par
Data will be made available on request.

~\\
\textbf{Acknowledgements}\par
\ \par
This work was partly supported by the National Natural Science Foundation of China (Nos. 12371354 and 12161141003), the Science and Technology Commission of Shanghai Municipality, China (No.22JC1403600),the National Key R\&D Program of China under Grant No. 2022YFA1006400, and the Fundamental Research Funds for the Central Universities, China.

\appendix
\section{}
\textbf{a. Algebraic properties of directed graphs}

For a simple unweighted directed graph $ G = (V, E) $, where the adjacency matrix is denoted by $ A $, the out-degree matrix is $ D^{+} = \text{diag}(d^{+}) $, and the in-degree matrix is $ D^{-} = \text{diag}(d^{-}) $. When all out-degrees are non-zero, i.e., $ d^{+} $ has no zeros, then $ D^{+} $ is invertible. Let $ P = (D^{+})^{-1}A $ represent the probability transition matrix of a Markov process on the directed graph. $ P $ is a non-negative, non-symmetric matrix with zeros on its diagonal, but no zero rows. It is evident that $ P\mathbf{1} = \mathbf{1} $. Let $ \rho(P) $ denote the spectral radius of $ P $, and $ \phi $ be the left eigenvector corresponding to $ \rho(P) $, satisfying $ \sum_{v \in V} \phi(v) = 1 $, i.e., $ \phi P = \rho(P)\phi $. Consequently, $ \phi P\mathbf{1} = \rho(P)\phi\mathbf{1} = \phi\mathbf{1} $, implying $ \rho=1 $, thus $ \phi P=\phi $.
\begin{lemma} \cite{horn2012matrix}\label{1.1}
    Let $A=[a_{ij}]\in M_{n}$ be non-negative. Then, $\rho\left(A\right)\leq|||A|||_{\infty}=max_{1\leq i\leq n}\sum_{j=1}^{n}a_{ij}$, and $\rho\left(A\right)\leq|||A|||_{1}=max_{1\leq j\leq n}\sum_{i=1}^{n}a_{ij}$; If all row sums of $A$ are equal, then $\rho\left(A\right)=|||A|||_{\infty}$; If all column sums of $A$ are equal, then $\rho\left(A\right)=|||A|||_{1}$.
\end{lemma}

\begin{theorem} \cite{horn2012matrix}\label{1.2}
    If $A\in M_{n}$ is non-negative, then $\rho\left(A\right)$ is an eigenvalue of $A$, and there exists a non-negative non-zero vector $x$, such that $Ax=\rho\left(A\right)x$.
\end{theorem}

\begin{theorem}\label{1.3}
    For a weakly connected but not strongly connected directed graph $G=(V,E)$, with all its strongly connected components being single nodes, there must exist a node $v$ with out-degree 0, and a node $w$ with in-degree 0.
\end{theorem}

\begin{proof}
    First, we prove the existence of a node with out-degree 0. Suppose all nodes of the directed graph $G$ have non-zero out-degrees. Starting from any node $v_{1}$ in $G$, continually find the out-neighboring nodes along the direction of edges, obtaining a directed path $P=\left\{v_{1},v_{2},...,v_{k}\right\}$ with distinct nodes, where $k\leq n$. It is known that the out-degree of node $v_{k}$ is non-zero. If there exists an out-neighboring node $v_{k+1}\in\left\{v_{1},v_{2},...,v_{k-1}\right\}$ for $v_{k}$, then a directed Hamiltonian cycle is formed, which is strongly connected. This contradicts the fact that all strongly connected components are single nodes. Therefore, all out-neighboring nodes of $v_{k}$ are "new", i.e., $v_{k+1}\in V-\left\{v_{1},v_{2},...,v_{k}\right\}$. Continuing this process, since the number of nodes in directed graph $G$ is finite, when $k=|V|=n$, $v_{k+1}\in V-\left\{v_{1},v_{2},...,v_{k}\right\}=\emptyset$, implying that $v_{n}$ has no "new" out-neighboring nodes, which contradicts the assumption that all nodes have non-zero out-degrees.

    The proof of the existence of a node with in-degree 0 is similar, requiring the continual search for "new" in-neighboring nodes from any node.
\end{proof}

\textbf{b. Necessary and sufficient conditions for strong connectivity in directed graphs}

\begin{theorem}
    For a weakly connected directed graph $G=(V,E)$ with non-zero out-degree vector $d^{+}$, $G$ is strongly connected if and only if all vectors $\phi$ have no zero components, where $\phi$ is the left eigenvector corresponding to the spectral radius of the matrix $P=(D^{+})^{-1}A$.
\end{theorem}

\begin{proof}
    ~\\
    \textbf{Necessity:} When $G$ is a strongly connected directed graph, its adjacency matrix $A$ is a non-negative irreducible matrix. Since all out-degrees of $G$ are non-zero, i.e., $D^{+}$ is invertible, then $P=(D^{+})^{-1}A$ is also a non-negative irreducible matrix. The transpose of $P$, denoted as $P^{T}$, is also non-negative irreducible. According to the Perron-Frobenius theorem, $P^{T}$ has a unique left eigenvector $\phi^{T}$ corresponding to $\rho\left(P\right)>0$ with $\sum_{v\in V}\phi^{T}\left(v\right)=1$, such that $P^{T}\phi^{T}=\rho\left(P\right)\phi^{T}$, where $\phi^{T}>0$. Hence, there exists a unique $\phi>0$ such that $\phi P=\rho\left(P\right)\phi$ and $\sum_{v\in V}\phi\left(v\right)=1$.
    ~\\
    \textbf{Sufficiency:} (By Contradiction) Suppose $G$ is a weakly connected directed graph with all out-degrees greater than 0 (but not strongly connected). We aim to prove that there exist components of $\phi$ as defined above that are zero. Clearly, since $P=(D^{+})^{-1}A$ and $P^{T}$ are non-negative matrices, by Theorem \ref{1.2}, there exists a non-negative non-zero vector $\phi$ such that $\phi P=\rho\left(P\right)\phi$. By Lemma \ref{1.1} and $P\mathbf{1}=\mathbf{1}$, we know that $\rho\left(P\right)=1$, hence $\phi P=\phi$.

    Given that the directed graph $G$ is not strongly connected, by Theorem \ref{1.3}, there must exist a strongly connected component $A\subseteq V$ with only inward edges, denoted by $\overline{A}=V-A$. For $\forall j\in A$,
    \begin{equation*}
        \phi_{j}=\sum_{i\in V,i\rightarrow{j}}\phi_{i}P_{ij}=\sum_{i\in A,i\rightarrow{j}}\phi_{i}P_{ij}+\sum_{k\in \overline{A},k\rightarrow{j}}\phi_{k}P_{kj},
    \end{equation*}
    Summing over $j\in A$,
    \begin{equation*}
        \sum_{j\in A}\phi_{j}=\sum_{j\in A}\sum_{i\in A,i\rightarrow{j}}\phi_{i}P_{ij}+\sum_{j\in A}\sum_{k\in \overline{A},k\rightarrow{j}}\phi_{k}P_{kj},
    \end{equation*}
    where
    \begin{equation*}
        \sum_{j\in A}\sum_{i\in A,i\rightarrow{j}}\phi_{i}P_{ij}=\sum_{i\in A}\sum_{j\in A,i\rightarrow{j}}\phi_{i}P_{ij}=\sum_{i\in A}\phi_{i},
    \end{equation*}
    Thus, $\sum_{j\in A}\sum_{k\in \overline{A},k\rightarrow{j}}\phi_{k}P_{kj}=0$. Since $P$ is defined such that $k\rightarrow j$ implies $P_{kj}>0$, for $\forall k\rightarrow j,j\in A , k\in\overline{A}$, we have $\phi_{k}=0$. Since $G$ is weakly connected, such $k$ must exist, implying that $\phi$ has zero components.
\end{proof}

From the above theorem, it is evident that for a weakly connected directed graph $G=(V,E)$ with non-zero out-degree vector $d^{+}$, $G$ is not strongly connected if and only if all vectors $\phi$ have zero components, where $\phi$ is the left eigenvector corresponding to the spectral radius of the matrix $P=(D^{+})^{-1}A$.

It is known that under the condition where $P$ is non-negative and non-symmetric (not fully positive and irreducible), $\phi$ may not be uniquely determined. However, the following theorem illustrates that all $\phi$ vectors can reflect the structural properties of a non-strongly connected graph, thus emphasizing the importance of $\phi$ in non-strongly connected graphs.
\begin{theorem}
    For a directed graph $G$ and vector $\phi$ as defined above, the non-zero elements in $\phi$ correspond to the nodes in $G$ that belong to strongly connected components with only inward edges, and the number of linearly independent $\phi$ equals the number of such strongly connected components in $G$.
\end{theorem}
\begin{proof}
    By Theorem \ref{1.3}, it is known that a non-strongly connected graph $G$ must have a strongly connected component $B$ with only outward edges. Let $\overline{B}=V-B$ denote the complement set of nodes. For $\forall j\in B$,
    \begin{equation*}
        \phi_{j}=\sum_{i\in B,i\rightarrow j}\phi_{i}P_{ij}=\sum_{k\in B,j\rightarrow k}\phi_{j}P_{jk}+\sum_{k\in \overline{B},j\rightarrow k}\phi_{j}P_{jk}
    \end{equation*}
    Summing over $j\in B$,
    \begin{equation*}
        \sum_{j\in B}\phi_{j}=\sum_{i,j\in B,i\rightarrow j}\phi_{i}P_{ij}=\sum_{j,k\in B,j\rightarrow k}\phi_{j}P_{jk}+\sum_{j\in B,k\in \overline{B},j\rightarrow k}\phi_{j}P_{jk}
    \end{equation*}
    Hence, $\sum_{j\in B,k\in \overline{B},j\rightarrow k}\phi_{j}P_{jk}=0$, so for $\forall\tilde{j}\in B$, when there exists $k\in\overline{B},\tilde{j}\rightarrow k$, $\phi_{\tilde{j}}=0$. Consequently, $\sum_{i\in B,i\rightarrow \tilde{j}}\phi_{i}P_{i\tilde{j}}=0$, i.e., for $\forall\Tilde{i}\in B$, when there exists the above $\tilde{j},\Tilde{i}\rightarrow\tilde{j}$, $\phi_{\Tilde{i}}=0$. This implies $\sum_{t\in B,t\rightarrow \tilde{i}}\phi_{t}P_{t\tilde{i}}=0$ for $\forall\Tilde{t}\in B$ when there exists the above $\tilde{i},\Tilde{t}\rightarrow\tilde{i}$. Since the component $B$ is strongly connected, repeating the above process yields $\forall j\in B,\phi_{j}=0$. Removing all components satisfying the conditions of $B$ in the non-strongly connected graph $G$ will result in new components satisfying the conditions of $B$. Eventually, $G$ will only consist of disconnected components, where the edges only node towards them. Therefore, the non-zero elements of $\phi$ must belong to these components.

    Since the initial state is arbitrary, the number of linearly independent $\phi$ equals the number of strongly connected components in $G$ with only inward edges, i.e., the number of remaining disconnected components at the end.
\end{proof}

\textbf{c. Algorithm for finding strongly connected components and eigenvector centrality on directed graphs}

\begin{algorithm}
	\renewcommand{\algorithmicrequire}{\textbf{Input:}}
	\renewcommand{\algorithmicensure}{\textbf{Output:}}
	\caption{Strongly Connected Components and Eigenvector Centrality}
	\label{alg1}
	\begin{algorithmic}[1]
        \REQUIRE Adjacency matrix A.
        \ENSURE  Strongly connected components $C_{i},i=1,2,...$;eigenvector centrality $f(x)$ on $C_{i}$.
		\WHILE{A is not zero}
		\STATE  $(D^{+})_{ii}=\sum_{j=1}^{n}A_{ij},P=(D^{+})^{-1}A$,
        \STATE  Find all vectors $\phi$ satisfying $\phi P=\phi,\sum_{v}\phi(v)=1$,
        \STATE  All connected components in $\phi$ with nonzero coordinates are $C_{i}$ in A,
		\STATE  Nonzero value of $\phi$ is equal to $f(x)$ of the node $x$ on $C_{i}$,
		\STATE  Update A by deleting nonzero nodes in $\phi$.
        \ENDWHILE
	\end{algorithmic}
\end{algorithm}

\section{}
The optimization problems (\ref{16}), (\ref{17}), and (\ref{18}) for solving the control variables $H$ and $F^{i}$ are equivalent \cite{nie2018multiview}.
\begin{proof}
~\\
$\bullet$ First, we prove $(\ref{16}) \iff (\ref{17})$.\\
Since $\sum_{i=1}^{v}p_{i}=1$, we have
\begin{equation*}
    \sum_{i=1}^{v}\frac{1}{p_{i}}||\overline{S}^{i}\tilde{S}^{i}-H(F^{i})^{T}||_{F}^{2}=(\sum_{i=1}^{v}\frac{1}{p_{i}}||\overline{S}^{i}\tilde{S}^{i}-H(F^{i})^{T}||_{F}^{2})(\sum_{i=1}^{v}p_{i})\geq(\sum_{i=1}^{v}||\overline{S}^{i}\tilde{S}^{i}-H(F^{i})^{T}||_{F})^{2}
\end{equation*}
where the inequality follows from the Cauchy inequality. It is easy to see that
\begin{equation*}
    (\sum_{i=1}^{v}||\overline{S}^{i}\tilde{S}^{i}-H(F^{i})^{T}||_{F})^{2}=\underset{p_{i}}{\text{min}} \quad\sum_{i=1}^{v}\frac{1}{p_{i}}||\overline{S}^{i}\tilde{S}^{i}-H(F^{i})^{T}||_{F}^{2}
\end{equation*}
Hence, we have
\begin{equation*}
    \begin{aligned}
         \quad &\underset{H,F^{i}}{\text{min}}\quad \sum_{i=1}^{v}||\overline{S}^{i}\Tilde{S}^{i}-H(F^{i})^{T}||_{F}\\
        \iff \quad &\underset{H,F^{i}}{\text{min}} \quad(\sum_{i=1}^{v}||\overline{S}^{i}\Tilde{S}^{i}-H(F^{i})^{T}||_{F})^{2}\\
        \iff \quad &\underset{H,F^{i},p_{i}}{\text{min}} \quad\sum_{i=1}^{v}\frac{1}{p_{i}}||\overline{S}^{i}\Tilde{S}^{i}-H(F^{i})^{T}||_{F}^{2}.
    \end{aligned}
\end{equation*}
~\\
$\bullet$ Next, we prove $(\ref{17}) \iff (\ref{18})$.\\
Let $H_{*}$ and $F_{*}^{i}$ be the optimal solution for problem (\ref{18}). We will now prove that they are also optimal for problem (\ref{17}). If the minimum value of the objective function in problem (\ref{17}) is greater than the minimum value of problem (\ref{18}), then obviously choosing $H_{*}$ and $F_{*}^{i}$ for problem (\ref{17}) would yield a smaller result. If problem (\ref{17}) has another optimal solution $H_{**}$ and $F_{**}^{i}$,
\begin{equation*}
    \begin{aligned}
        \quad & \sum_{i=1}^{v}\frac{1}{||\overline{S}^{i}\Tilde{S}^{i}-H_{**}(F_{**}^{i})^{T}||_{F}}||\overline{S}^{i}\Tilde{S}^{i}-H_{**}(F_{**}^{i})^{T}||_{F}^{2}\\
         =\quad &\sum_{i=1}^{v}||\overline{S}^{i}\Tilde{S}^{i}-H_{**}(F_{**}^{i})^{T}||_{F}\\
        <\quad & \sum_{i=1}^{v}||\overline{S}^{i}\Tilde{S}^{i}-H_{*}(F_{*}^{i})^{T}||_{F}\\
        =\quad & \sum_{i=1}^{v}\frac{1}{||\overline{S}^{i}\Tilde{S}^{i}-H_{*}(F_{*}^{i})^{T}||_{F}}||\overline{S}^{i}\Tilde{S}^{i}-H_{*}(F_{*}^{i})^{T}||_{F}^{2},
    \end{aligned}
\end{equation*}
Therefore, $H_{*}$ and $F_{*}^{i}$ are not the optimal solutions for problem (\ref{18}), which contradicts the assumption.
\end{proof}


\bibliographystyle{elsarticle-num}
\bibliography{references}






\end{document}